\documentclass[preprint,12pt,authoryear]{elsarticle}

\usepackage{natbib}
\usepackage{amsmath,amsfonts,amssymb,mathtools}
\usepackage{commath}
\usepackage{bm}
\usepackage{lipsum}
\usepackage{multicol}
\usepackage{pdfpages}
\usepackage{epstopdf}
\usepackage{multirow}
\usepackage{booktabs}
\usepackage{afterpage}
\usepackage{soul}
\usepackage{pifont}

\usepackage{hyperref}
\usepackage{color,soul}
\usepackage{array}
\usepackage{float}
\usepackage{svg}
\usepackage{subcaption}
\newcolumntype{L}{>{\centering\arraybackslash}m{3cm}}
\usepackage{xcolor}
\hypersetup{
     colorlinks = true,
     linkcolor = blue,
     anchorcolor = blue,
     citecolor = green,
     filecolor = magenta,
     urlcolor = cyan
     }

\usepackage{titlesec}
\titleformat{\subsection}
  {\normalfont\bfseries}{\thesubsection}{1em}{}
\titleformat{\subsubsection}
  {\normalfont\bfseries}{\thesubsection}{1em}{}

\begin{document}

\begin{frontmatter}



\title{Exploring Contextual Representation and Multi-Modality for End-to-End Autonomous Driving}


\author[inst1,inst2]{Shoaib Azam}
\author[inst1,inst2]{Farzeen Munir}
\author[inst1]{Ville Kyrki}
\author[inst3]{Moongu Jeon}
\author[inst4,inst5,inst6]{Witold Pedrycz}

\affiliation[inst1]{organization={Department of Electrical Engineering and Automation, Aalto University, Espoo, Finland}
            }
\affiliation[inst2]{organization={Finnish Center for Artificial Intelligence, Finland}
            }

\affiliation[inst3]{organization={School of Electrical Engineering and Computer Science, Gwangju Institute of Science and Technology, Gwangju, 61005, South Korea }
}
            
\affiliation[inst4]{organization={Department of Electrical and Computer Engineering},
            addressline={University of Alberta}, 
            city={Edmonton},
            postcode={AB T6R 2V4}, 
            country={Canada}}

\affiliation[inst5]{organization={Department of Electrical and Computer Engineering, Faculty of Engineering},
            addressline={King Abdulaziz University}, 
            city={Jeddah},
            postcode={21589}, 
            country={Saudia Arabia}}

\affiliation[inst6]{organization={Systems Research Institute},
            addressline={Polish Academy of Sciences}, 
            city={Warsaw},
            postcode={01-447}, 
            country={Poland}}

\begin{abstract}
{Learning contextual and spatial environmental representations enhances autonomous vehicle's hazard anticipation and decision-making in complex scenarios. Recent perception systems enhance spatial understanding with sensor fusion but often lack full environmental context. Humans, when driving, naturally employ neural maps that integrate various factors such as historical data, situational subtleties, and behavioral predictions of other road users to form a rich contextual understanding of their surroundings. This neural map-based comprehension is integral to making informed decisions on the road. In contrast, even with their significant advancements, autonomous systems have yet to fully harness this depth of human-like contextual understanding. Motivated by this, our work draws inspiration from human driving patterns and seeks to formalize the sensor fusion approach within an end-to-end autonomous driving framework. We introduce a framework that integrates three cameras (left, right, and center) to emulate the human field of view, coupled with top-down bird-eye-view semantic data to enhance contextual representation. The sensor data is fused and encoded using a self-attention mechanism, leading to an auto-regressive waypoint prediction module. We treat feature representation as a sequential problem, employing a vision transformer to distill the contextual interplay between sensor modalities. The efficacy of the proposed method is experimentally evaluated in both open and closed-loop settings. Our method achieves displacement error by $0.67m$ in open-loop settings, surpassing current methods by $6.9\%$ on the nuScenes dataset. In closed-loop evaluations on CARLA's Town05 Long and Longest6 benchmarks, the proposed method enhances driving performance, route completion, and reduces infractions.
}
\end{abstract}




\end{frontmatter}



\section{Introduction}\label{sec1}
The ecosystem of autonomous driving involves the perception and planning modules to complement each other for a smooth course of action \citep{yurtsever2020survey}. To this end, two approaches---modular \citep{azam2020system} and end-to-end autonomous driving \citep{xiao2020multimodal,khan2022level}---have been adopted in academia and industry as possible solutions for perception and planning modules. Although both approaches have their pros and cons, on the level of scalability, end-to-end autonomous driving is more promising in contrast to the modular approach. End-to-end autonomous driving bridges the perception and planning in a unified framework providing a differentiable stack for learning the driving policies \citep{schwarting2018planning}. 
\par 
Extracting meaningful spatial and temporal scene representation is vital for learning better driving policies in end-to-end autonomous driving. In the literature, various sensor modalities, such as single-camera systems and lidar, have been employed to capture environmental details. Techniques have been developed to extract both spatial and temporal information from these modalities \citep{huang2020multi,behl2020label}. Among these, sensor fusion techniques stand out, integrating data from multiple sensors to achieve a comprehensive understanding. However, obtaining a complete understanding of a complex and dynamic environment—which involves interactions between multiple dynamic agents and features across different views or modalities—necessitates a global context. This global context aids in establishing a contextual representation across different modalities.
\par

This work explores the formulation of driving policies by integrating multi-camera perspectives with top-down bird-eye-view (BEV) semantic maps to provide a holistic view of the environment. This integrative technique aims to refine the perception capabilities within an end-to-end autonomous driving framework. Our framework initiates by extracting and then fusing features from the input data, applying a transformer network to correlate these features with the subsequent control commands for the vehicle. The autonomous vehicle's waypoints are informed by the insights gained from the transformer. To validate our proposed method, we have conducted evaluations in both open-loop and closed-loop contexts, employing the nuScenes dataset and the Town05 Long and Longest6 Carla benchmarks, respectively. The results indicate that our approach outperforms existing state-of-the-art methods in the precision of driving policy prediction.
\par
The main contribution of this work are:
\begin{enumerate}
\item Designing a framework that demonstrates an integration of spatial perception through RGB cameras with a top-down bird's-eye view (BEV) for contextual mapping. This dual approach mimics human-like perception by combining immediate visual data with a global understanding of the environment, enhancing the autonomous system's ability to navigate complex scenarios
\item Develop a transformer-based encoder to sequence the spatial and contextual features, leading to an improved feature representation for learning the driving policies.

\end{enumerate}
\par
The rest of the paper is organized as follows: Section \ref{Related} covers the literature review. Section \ref{Method} presents the proposed  framework and Section \ref{Experiments} presents experimentation, analysis and results. Finally, Section \ref{Conclusion} concludes the paper with possible directions of future work.

\section{Related Work} \label{Related}
\subsection{\textbf{Multi-modal End-to-end Learning Frameworks for Autonomous Driving}}
Learning optimal trajectories involve a better representation of the environment to include spatial, temporal, and contextual information of the environment. Different multi-modal end-to-end driving methods are developed in the literature to improve driving performance. These multi-modal methods either use cameras, Lidar, HD maps, or sensor fusion between these information modalities. \cite{xiao2020multimodal} have used the sensor fusion between RGB cameras and depth information to investigate the use of multi-modal data compared to single modality for end-to-end autonomous driving. Some works have focused on semantics and depth for determining the explicit intermediate representation of the environment and their effect on autonomous driving \citep{behl2020label,zhou2019does}. In addition, some works, for instance, NMP \citep{zeng2019end}, have used the Lidar and HD maps first to generate the intermediate $3$D detections of the actors in the future and then learn a cost volume for choosing the best trajectory. Lidar and camera fusion are extensively used for perception and obtaining driving policies. \cite{sobh2018end} have used the Lidar and image fusion by processing both sensor modality streams in a separate branch and then fusing the resulting features. Further, they have applied semantic segmentation and Lidar post-processing Post Grid Mapping to increase the method's robustness. Similarly, \cite{prakash2021multi} have fused the Lidar and camera data at multiple levels through self-attention for learning the driving policies. In addition, some methods have adopted sensor fusion between camera and semantic maps \citep{natan2022end} for learning end-to-end driving policy for autonomous driving. Several studies have investigated the application of knowledge distillation techniques to learn driving policies. In this approach, a privileged agent is initially trained with access to comprehensive information, such as maps, navigational data, and images. Subsequently, this privileged agent is employed to train a sensorimotor agent, which only has access to image data \citep{chen2020learning,zhang2023coaching}. Furthermore, improving the decoder architecture in an encoder-decoder architecture is also being explored by \citep{jia2023think}. 
All these methods have used sensor fusion techniques to acquire the spatial or temporal information of the environment but lack contextual information in terms of BEV semantic maps. In the proposed work, we have opted for BEV semantic maps and incorporated them with a camera stream to answer whether the inclusion of BEV semantic maps improves driving performance. 
\subsection{\textbf{BEV Representation End-to-end Autonomous Driving}}
Representing the environment in a BEV benefits the planning and control task as it circumvents the issues like occlusion and scale distortion and also provides the contextual representation of the environment. In this context, some works focus on generating the BEV representation; for instance, ST-P3 leverages spatial-temporal learning by designing an egocentric-aligned representation of BEV and finally uses that representation for perception, planning, and control \citep{hu2022st}. \cite{hu2023planning} have designed an end-to-end planning  autonomous driving framework. This framework's perception and prediction modules are structured as transformer decoders, with task queries acting as the interface between these two nodes. An attention-based planner is used to sample the future waypoints by considering the past node's data. Following the same approach, \cite{jiang2023vad} have used a vectorized representation for end-to-end autonomous driving. They have adopted a BEV encoder for BEV feature extraction combined with map and agent queries in a transformer network for environment representation and then a planning transformer for predicting the trajectories. In addition, \cite{chitta2021neat} have proposed a neural attention field for waypoint prediction. All these methods have used the BEV representation, similar to our work, but are more focused on how to make the BEV representation from input images; however, in our work, we focus on how to use the BEV features for learning the policy rather make BEV from input images and then use it for the planning. The experimental analysis shows the efficacy of our proposed method against state-of-the-art methods illustrating the effectiveness of using BEV representation for learning the driving policies in both open and closed-loop settings.
\subsection{Transformer in End-to-End Autonomous driving}
Initially used for natural language processing tasks \citep{vaswani2017attention}, transformers have widely been employed for learning meaningful representation in vision applications \citep{dosovitskiy2020image,carion2020end}. The transformer's self-attention module enhances the learning of sequential data globally and improves feature representation. \cite{prakash2021multi} employed the transformer to combine intermediate features representation from RGB images and Lidar data. \cite{shao2022safety} proposed an encoder-decoder transformer network for predicting the waypoints using fusion between different view-point cameras and Lidar to learn the global context for comprehensive scene understanding. \cite{huang2022multi} design a transformer-based neural prediction framework that considers social interactions between different agents and generates possible trajectories for autonomous vehicles. \cite{dong2021image} determines the driving direction from visual features acquired from images by using a novel framework consisting of a visual transformer. The driving directions are decoded for human interpretability to provide insight into learned features of the framework. Finally, \cite{li2020end} considers social interaction between agents on the road and forecasts their future motion. The spatial-temporal dependencies were captured using a recurrent neural network combined with a transformer encoder.

\begin{figure*}[t]
      \centering
      \includegraphics[width=\textwidth, keepaspectratio]{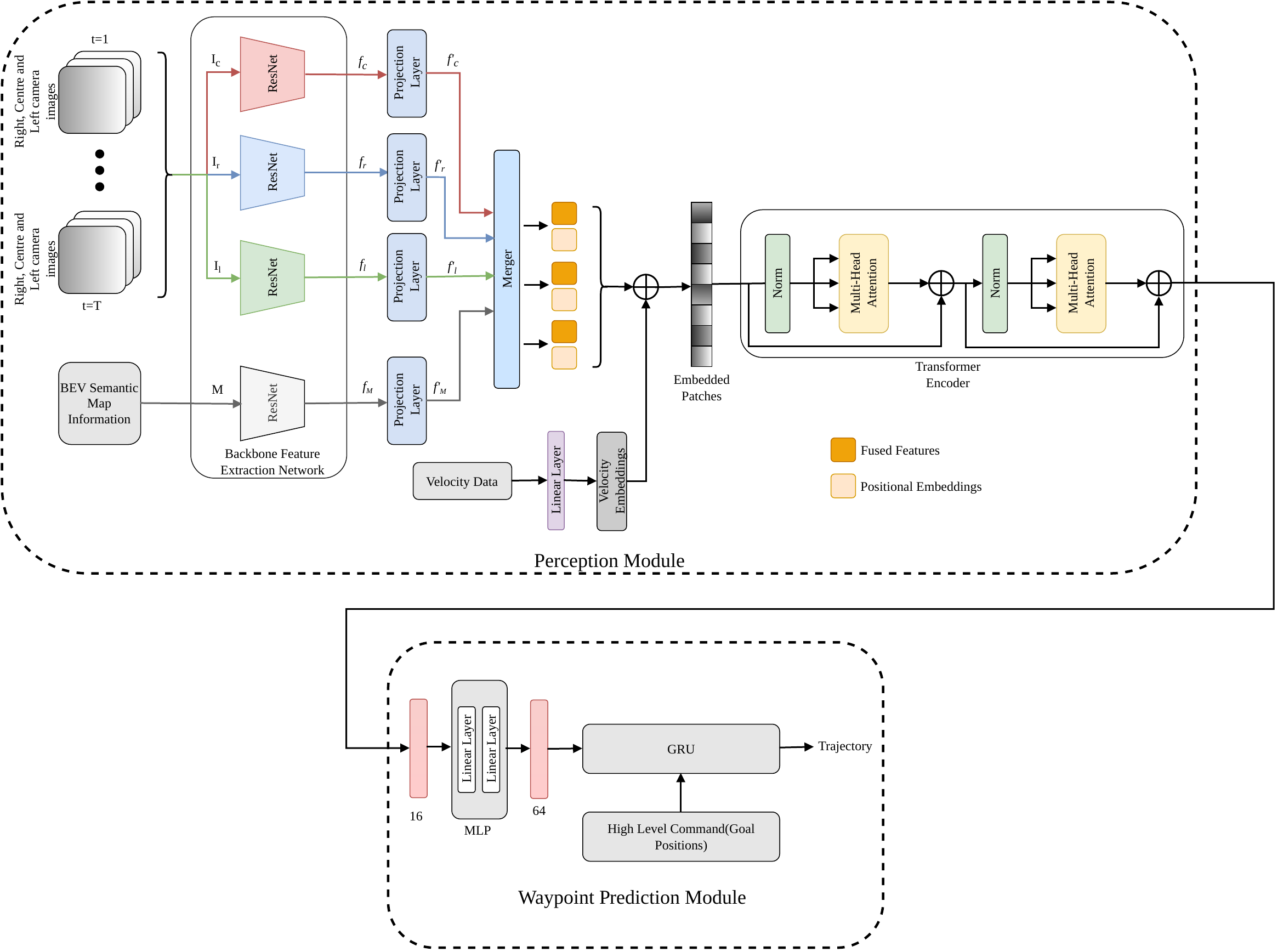}
      \caption{The architecture of the proposed method which is comprised of two modules: perception block and waypoint prediction block. The perception module generates the features extracted from the input three RGB cameras (center, left, right) and the top-down semantic maps. These extracted features are then embedded with the velocity information to be utilized by the transformer encoder. The encoded features are then passed to the GRU-based waypoint prediction module for the generation of next waypoints. (Best view in color)}
      \label{framework}
\end{figure*}

\section{Method} \label{Method}
An overview of the proposed method is depicted in Figure \ref{framework}. This method encompasses both a perception module and a waypoint prediction module, collectively facilitating the learning of end-to-end driving policies. The perception module extracts feature from input sensor modalities and then passes them to the waypoint prediction module to generate future waypoints/trajectories. The following sections detail the problem formulation, and model architecture, explaining the perception and waypoint prediction modules. 

\subsection{Problem Formulation}
In this work, an end-to-end learning approach is adopted for the point-to-point navigation problem, where the objective of the trained agent is to safely reach the goal point by learning a driving policy ${\pi^*}$ that imitates the expert policy $\pi$. The learned policy completes the given route by avoiding obstacles and complying with the traffic rules. In the closed-loop settings, we have opted for the CARLA simulator to collect the expert dataset in a supervised learning approach. Similarly, to use the expert data in open-loop settings, we have used the nuScenes dataset. 
Suppose the dataset $D = {(X^j, Y^j)}_{j=1}^{d}$ of size $d$ is collected that consists of high dimensional observations vector $X$ from the sensory modalities along with the corresponding expert trajectories vector $Y$. The expert trajectories are defined in vehicle local coordinate space and are set of $2$D waypoints transformed that is, $Y = {\mathbf{y_t} = (u_t,v_t)_{t=1}^{T}}$, where $u_t$ and $v_t$ are the position information in horizontal and vertical directions, and $T$ corresponds to the future horizon for the waypoints, respectively. The objective is to learn the policy $\pi$ with the collected dataset $D$ in a supervised learning framework with the loss function $\mathcal{L}$ expressed as follows
\begin{equation}
\label{equ1}
\arg \min _{\pi}\mathbb{E}_{(X,Y)\rightarrow D}[\mathcal{L}(Y,\pi(X))].
\end{equation}
In this urban setting, the high-dimensional observations include the center, right, left cameras and top-down BEV semantic data. 

\subsection{Model Architecture}
This section explains the proposed method model architecture which is composed of two main modules i) perception and ii)waypoint prediction module. The perception module accepts the sensor modalities information for the feature extraction which is then passed to the waypoint prediction module for the prediction of waypoints using an auto-regressive model.
\subsubsection{Perception Module}
The perception module of the proposed method includes the backbone architecture and transformer encoder network. The following subsections explain the details of the perception module.
\newline
\textbf{Backbone}:
In this module of the proposed method, we have to build a spatiotemporal representation of the environment. 
The sequence of input RGB images from all three views, that is, center ($I_c  {\in \mathcal{R}^{3 \times H \times W}}$), right ($I_r  {\in \mathcal{R}^{3 \times H \times W}}$), and left ($I_l  {\in \mathcal{R}^{3 \times H \times W}}$) having width $W$ and height $H$ are processed by the backbone network for the feature extraction. ResNet architecture is employed for feature extraction in the proposed method, as it is efficient  for feature representation. Similarly, the BEV semantic maps $(M {\in \mathcal{R}^{H \times W}})$ are passed to the ResNet network for feature extraction. In the case of closed-loop settings, these BEV semantic maps are collected through the simulator, whereas in the case of open-loop settings, the BEV semantic maps are generated from the nuScenes dataset, as explained in Section $3.4.2$.
In our settings, the pre-trained ResNet model, trained on the ImageNet dataset, is utilized for generating the low-resolution feature maps $f {\in \mathcal{R}^{C \times H \times W}}$ for each sensor modality. The last layer of the ResNet model for each sensor modality generates ($B,512,8,8$) dimension feature maps, where $B$ corresponds to batch size. The resolution of these resulting feature maps is reduced to the dimension of ($B,512,1,1$) by average pooling and flattened to a $512$ dimensional vector. To input the features into the transformer encoder, a projection layer is used to transform the $512$ dimensional vector into a $400$ dimensional vector.
Finally, all the $400$ dimensional vectors from each sensor modality that includes center $f'_c$, right $f'_r$, left $f'_l$ cameras, and top-down BEV semantic maps $f'_M$ are concatenated to give the final $1600$ dimensional vector, which is then reshaped to ($B,1,40,40$) to be utilized by the transformer encoder. The following expressions in (\ref{equ2}) summarize the computation of features maps in the backbone network for each sensor modality and top-down semantic maps,
\begin{equation}
\small
\label{equ2}
\begin{aligned}
& I_{c,r,l}^{\mathcal{R}^{3 \times H \times W}} \xRightarrow{\text{Conv, bn, relu}} f_{c,r,l}^{\mathcal{R}^{64 \times 128 \times 128}}, \\
& f_{c,r,l}^{\mathcal{R}^{64 \times 128 \times 128}} \xRightarrow{\text{maxpool}} f_{c,r,l}^{\mathcal{R}^{64 \times 64 \times 64}}, \\
& f_{c,r,l}^{\mathcal{R}^{64 \times 64 \times 64}} \xRightarrow{l_1} f_{c,r,l}^{\mathcal{R}^{64 \times 64 \times 64}}, \\
& f_{c,r,l}^{\mathcal{R}^{64 \times 64 \times 64}} \xRightarrow{l_2} f_{c,r,l}^{\mathcal{R}^{128 \times 32 \times 32}}, \\
& f_{c,r,l}^{\mathcal{R}^{128 \times 32 \times 32}} \xRightarrow{l_3} f_{c,r,l}^{\mathcal{R}^{256 \times 16 \times 16}}, \\
& f_{c,r,l}^{\mathcal{R}^{256 \times 16 \times 16}} \xRightarrow{l_4} f_{c,r,l}^{\mathcal{R}^{512 \times 8 \times 8}}, \\
& f_{c,r,l}^{\mathcal{R}^{512 \times 8 \times 8}} \xRightarrow{\text{avgpool}} f_{c,r,l}^{\mathcal{R}^{512 \times 1 \times 1}}, \\
& f_{c,r,l}^{\mathcal{R}^{512 \times 1 \times 1}} \xRightarrow{\text{flatten}} f_{c,r,l}^{\mathcal{R}^{512}}, \\
& f_{c,r,l}^{\mathcal{R}^{512}} \xRightarrow{\text{proj}} f_{c,r,l}^{\mathcal{R}^{400}} \xRightarrow{\text{reshape}} f_{c,r,l}^{\mathcal{R}^{1,40,40}},
\end{aligned}
\end{equation}
where $H,W = 256$ respectively for each sensor modality and top-down BEV semantic maps. The expressions mentioned above only illustrate the computation of feature extraction for a single batch. In our experiments, we have used the batch size of $64$ to train the proposed method.\\
\newline
\textbf{Transformer Encoder}:
In this work, a transformer encoder, specifically a vision transformer, is employed to learn the contextual relationship between the features and to generalize it to learn better feature representation. In this context, the resulting features $f = \mathcal{R}^{1 \times H \times W}$ is fed to the transformer encoder by flattening into patches $f_p = \mathcal{R}^{N \times (P^2C)}$, where $H$ and $W$ corresponds to the resolution of input features from the backbone network, $C$ is the number of channels, ($P, P$) is the size of each patch, and $N=HW/P$ denotes the number of patches and also the input sequence length. In addition, a learnable position embedding is added to the input sequence, a trainable parameter with the same dimension as the input sequence, so that the network infers the spatial dependencies between different tokens at the train time. A velocity embedding is also added to the $C$ dimensional of the input sequence through a linear layer, which includes the current velocity. Finally, the input sequence, positional embeddings $ E_{pos}$, and velocity embeddings $E_{vel}$ are element-wise summed together,which is mathematically expressed in the following,
\begin{equation}
\small
\label{equ3}
\begin{aligned}
& z_o = [f_p^1E;f_p^2E;...;f_p^NE] + E_{pos}+ E_{vel}, \\ 
& E \in \mathcal{R}^{(P^2.C) \times D}, \\
& E_{pos} \in \mathcal{R}^{(N+1) \times D}, E_{vel} \in \mathcal{R}^{(N+1) \times D}, \\
& {z}'_l = MSA(LN(z_{l-1}))+z_{l-1}+z_{l-1}, \\
& z_l = MLP(LN({z}'_l))+{z}'_l+{z}'_l,
\end{aligned}
\end{equation}
where MSA corresponds to multi-head self-attention, MLP is multi-layer perceptron, LN is layer normalization, and $D$ corresponds to dimension. The multi-head attention helps in generating the rich feature representation for the input sensor modalities that in turn to learn better contextual representation. The  formulation of the multi-head self-attention is expressed as,
\begin{equation}
\small
\label{equ4}
\begin{aligned}
& (\mathbf{Q},\mathbf{K},\mathbf{V}) = \mathbf{z}\mathbf{W}_{QKV}, \\
& \mathbf{W}_{QKV} \in \mathcal{R}^{D \times 3D_h}, \\
& A = softmax(\mathbf{Q}\mathbf{K^T})/\sqrt{D_h}, \\
& A \in \mathcal{R}^{N \times N}, \\
& SA(\mathbf{z}) = A\mathbf{v}, \\
& MSA(\mathbf{z}) = [SA_1(\mathbf{z};SA_2(\mathbf{z};...;SA_j(\mathbf{z})]\mathbf{W}_{msa}, \\
& \mathbf{W}_{msa} \in \mathcal{R}^{(j.D_h) \times D}.
\end{aligned}
\end{equation}
where $\mathbf{Q}$, $\mathbf{V}$ and $\mathbf{K}$ are the query, value and key vectors and $\mathbf{W}$ is the weight matrix. The output features from the MSA have the same dimensionality as the input features. The transformer encoder applies the attention multiple times throughout the architecture. The final output features from the transformer encoder are then summed along the dimension to produce the $16$ dimensional vector having the contextual representation of features from all the sensor modalities. This resulting $16$ dimensional feature vector is injected into the waypoint prediction module to predict waypoints.

\subsubsection{Waypoint Prediction Module}
The waypoint prediction module acts as a decoder for predicting future waypoints using the encoded information from the transformer encoder. The resulting $16$ dimensional vector is passed through an MLP consisting of two hidden layers having $256$ and $128$ units, respectively, to output the $64$ dimensional vector. The MLP layer is used for upsampling the vector dimension from $16$ to $64$ and is related to experimental heuristics that produce better results in terms of waypoint prediction. We have employed the auto-regressive GRU model to predict the next waypoints that take the $64$ dimension feature vector to initiate the hidden state of the GRU model. The GRU-based auto-regressive model takes the current position and goal location as high-level commands as input, which helps the network focus on the relevant context in the hidden states to predict the next waypoints. In the case of closed-loop settings, the goal locations include the GPS points registered in the same ego-vehicle coordinate frame as input to the GRU rather than the encoder because of the colinear BEV space between the predicted waypoints and the goal locations. However, high-level commands such as forward, turn right and left are passed as input to the GRU for waypoint predictions in the open-loop settings. 
\par
In the open-loop settings, we have evaluated the predicted trajectory with the ground-truth trajectory without using a controller. However, for the closed-loop setting, the predicted waypoints are passed to the control module of the CARLA simulator to generate steer, throttle, and brake values. Two PID controllers for lateral and longitudinal control are used in this context. The longitudinal controller takes the average weighted magnitude of vectors between the waypoints of consecutive time steps, whereas the lateral control takes their orientation. For the control settings, we have used the settings as suggested by the following codebase benchmarked on the CARLA dataset.

\section{Experiments} \label{Experiments}
This section explains the proposed method evaluation in both open-loop and closed-loop settings. The nuScenes dataset is utilized for the open-loop evaluation, whereas the CARLA simulator is used for the closed-loop evaluation. 
\subsection{\textbf{Open-loop Experiments on nuScenes}}
\noindent \textbf{Dataset:} The nuScenes dataset contains $1$k diverse scenes comprising different weather and traffic conditions. Each scene is $20$ seconds long and contains $40$ frames, corresponding to a total of $40$k samples in the dataset. The dataset is recorded using a camera rig comprised of $6$ cameras on ego-vehicle, giving a full $360\deg$ view of the environment. The dataset includes the calibrated intrinsic $K$ and extrinsic $(R,t)$ for each camera view at every time-step. The proposed method settings utilize the center, right, and left camera views. Since the nuScenes dataset does not provide any top-down BEV semantic representations, the BEV semantic representation is generated using ego-vehicle poses and camera views intrinsic and extrinsic calibration data. 
\newline
\textbf{Input Representations:} For the nuScenes dataset, the input image from the center, front, and left camera views are first cropped and resized to $256 \times 256$ from the original resolutions of $900 \times 1600$. Contrary to the camera views and ego-vehicle future positions data, the nuScenes data does not provide the top-down BEV semantic maps. Since the proposed method requires the BEV semantic maps, we have generated those map labels using the ego poses and camera calibration data. To do so, for each camera, the camera's intrinsic $K$ matrix and extrinsic $T_{C}^{V}$ camera-to-vehicle calibration matrix is obtained from the nuScenes dataset. Similarly, the ego pose transformation matrix $T_{V}^{W}$ vehicle-to-world, which includes rotation matrix $R$ and translation vector $t$, are retrieved from the dataset. The BEV transformation matrix $T_{W}^{BEV}$ is obtained by taking the product between $T_{C}^{V}$, $R$, and $T_{V}^{W}$. The $3$D point or object is projected by first representing the $3$D point or object as a homogeneous vector $P_w[X_w, Y_w, Z_w]^T$. Then, the $3$D point or object is projected onto the BEV space by taking the product between the BEV transformation matrix $T_{W}^{BEV}$ and the homogeneous vector $P_w[X_w, Y_w, Z_w]^T$, resulting in the $2$D BEV coordinate $P_{bev}$. Finally, the segmentation map assigns class labels to the BEV pixel coordinates, giving the final BEV semantic map$M$. In our settings, we kept the resolution of this BEV semantic map to $256 \times 256$. 
\newline
\textbf{Output Representations:} The proposed method predicts the future trajectory $Y$, for the ego-vehicle in the ego-vehicle coordinate. In the open-loop settings, the future trajectory $Y$ is represented as waypoints that include position information. In our experiments, by default, the horizon $T=2.0s$ is set for predicting the future trajectory by taking the past $1.0s$ past context. 
\newline
\textbf{Evaluation Metrics:} For the proposed method evaluation, \textbf{Euclidean distance (L$2$ error)} is used which is the measure of distance between the expert trajectory and the predicted trajectory. Mathematically, the L$2$ error is defined by the Eq.\ref{L2}
\begin{equation}
\small
\label{L2}
L2(T_e,T_p) = \sum_{i=1}^{n}\sum_{j=1}^{d}(T_{_e{_{ij}}} - T_{_p{_{ij}}})^2 ,
\end{equation}
where,  $T_e$ and $T_p$ correspond to the expert and predicted trajectory, respectively. Each trajectory consists of $n$ points in a $d$-dimensional space. 

\begin{table*}[t]
\centering
\caption{Dataset generation details using the CARLA simulator for the proposed method}
\label{table-1}
\resizebox{13cm}{!}{%
\begin{tabular}{ll}
\toprule
Maps & Training Data: Town01, Town02, Town03, Town04, Town06, Town07, Town10 \\
& Test Data: Town05 \\
\midrule
Weather Conditions & Clear sunset, Clear noon, Wet noon, Wet sunset, \\
& Cloudy noon, Cloudy sunset, Rainy noon, Rainy sunset \\
\midrule
Non-player characters (NPCs) & Pedestrians, Car, Bicycle, Truck, Motorbike \\
\midrule
Object Classes & 0:Unlabeled, 1:Pedestrian, 2:Road line, 3:Road, \\
& 4:Sidewalk, 5:Car, 6:Red traffic light, \\
& 7: Yellow traffic light, 8: Green traffic light \\
\midrule
Routes & Tiny (only straight or one turn) \\
& Short (100-500m) \\
& Long (1000-2000m) \\
\midrule
CARLA Version & 0.9.10 \\
\bottomrule
\end{tabular}
}
\end{table*}

\subsection{Closed-loop Experiments on CARLA}
\noindent \textbf{Dataset:}
In this work, CARLA 0.9.10 \footnote{\url{https://carla.org/}} simulator is used to create a dataset for training and evaluation. Table \ref{table-1} illustrates the dataset details that are utilized in generating the training dataset to create a more varying simulation environment. For generating the dataset, an expert policy with the privileged information from the simulation is rolled out to save the data at $2$FPS. The dataset includes left, right, and center camera RGB images, top-down semantic map information, the corresponding expert trajectory, speed data, and vehicular controls. The trajectory includes $2$D waypoints transformed into BEV space in the vehicle's local coordinate, whereas the steering, throttle, and brake data are incorporated into the vehicular control data at the time of recording. Inspired by \cite{prakash2021multi} configurations, we have gathered the data by giving a set of predefined routes to the expert in driving the ego-vehicle. The GPS coordinates define the routes provided by the global planner and high-level navigational commands (e.g., turn right, follow the lane, etc). We have generated around $60$ hours of the dataset, including $200K$ frames.
\newline
\textbf{Input Representation:} 
The proposed method utilizes two modalities: RGB cameras (left, center and right) and semantic maps. The three RGB cameras provide a complete field of view that mimics the human field of view. The semantic maps are converted to BEV representation that contains ground-truth lane information, location, and status of traffic lights, vehicles, and pedestrians in the vicinity of ego-vehicle. The top-down semantic maps are cropped to the resolution of $ 256\times 256$ pixels. For all three cameras, to cater the radial distortion, the resolution is cropped to $256 \times 256$ from the original camera's resolution of $400 \times 300$ pixels at the time of extracting the data.
\newline
\textbf{Output Representation:} 
For the point-to-point navigation task, the proposed method predicts the future trajectory $Y$ of the ego-vehicle in the vehicle coordinate space. The future trajectory $Y$ is represented by a sequence of $2D$ waypoints, $Y = {\mathbf{y_t} = (u_t,v_t)_{t=1}^{T}}$, where $u_t$ and $v_t$ are the position information in horizontal and vertical directions, respectively. In the experimental analysis, we have utilized $T=4$ as the number of waypoints.
\newline
\textbf{Evaluation Metrics:}
The proposed method's efficacy is evaluated using the following metrics indicated by the CARLA driving benchmarks.
\newline 
\indent \underline {Route Completion}: is the percentage of route distance $R_j$ completed by the agent in route $j$ averaged across the number of $N$ routes is shown in the form,

\begin{equation}
\small
\label{RC}
RC = \frac{1}{N}\sum_{j}^{N}R_j.
\end{equation}
The RC is reduced if the agent drives off the specified route by some percentage of the route. This reduction in RC is defined by a multiplier ($1$-\% off route distance). 

\indent \underline {Infraction Multiplier}: as shown in (\ref{IM}) is defined as the geometric series of infraction penalty coefficient, $p^i$, for every infraction encountered by the agent along the route. Initially, the agent starts with the ideal base score of $1.0$, which is reduced by a penalty coefficient for every infraction. The penalty coefficient $p^i$ for each infraction is predefined. If the agent collides with the pedestrian $p_{pedestrian}$, the penalty is set to $0.50$; with other vehicles $p_{vehicles},$ it is set to $0.60$, $0.65$ for collision with static layout $p_{stat}$, and $0.7$ if the agent breaks the red light $p_{red}$. The penalty coefficient is defined as $PC={p_{pedestrian}, p_{vehicles}, p_{stat}, p_{red}}$,
\begin{equation}
\small
\label{IM}
IM = \prod_{i}^{PC}(p^{i})^{infractions^{i}}.
\end{equation}

\indent \underline{Driving Score}: is computed by taking the product between the percentage of the route completed by the agent $R_j$ and the infraction multiplier ${IM}_j$of the route $j$ and averaged by the number of the routes $N_r$. Higher  driving score corresponds to the better model. Mathematically, the driving score (DS) is 
\begin{equation}
\small
\label{DS}
DS = \frac{1}{N_r}\sum_{j=1}^{N_r}{RC}_j{IM}_j(p^{i})^{infractions^{i}}.
\end{equation}
It is to be noted that if the ego vehicle deviates from the route $j$ for more than $30$ meters or there is no action for $180$ seconds, then the evaluation process on route $j$ will be stopped to save the computations cost and next route will be selected for the evaluation process.

\subsection{Training Details}

The proposed method is trained using the dataset collected from the CARLA simulator by rolling out the expert model and also on the nuScenes dataset. In addition, we have used the pre-trained ResNet model trained on the ImageNet dataset to extract the features in the backbone network for each sensor modality. In training the proposed network, we have added augmentation such as rotating and noise injection to the training data, along with adjusting the waypoints labels. For the transformer encoder, we have used the patch size of $4$, which gives the $16$ dimensional feature embedding. We have trained the proposed method using the Pytorch library on RTX $3090$ having $24$ GB GPU memory for a total of $100$ epochs. In training, we have used the batch size of $64$ and an initial learning rate of $10^{-4}$, which is reduced by a factor of $10$ after every 20 epochs. The $L_1$ loss function is used for training the proposed method. Let $y_t^{gt}$ represent the ground-truth waypoints from the expert for the timestep $t$; then the loss function is represented as
\begin{equation}
\small
\label{loss}
\begin{aligned}
\mathcal{L} = \sum_{t=1}^{T}\left \| y_t - y_t^{gt} \right \|_1.
 \end{aligned}
\end{equation}
An AdamW optimizer  is used in training with a weight decay set to $0.01$ and beta values to the Pytorch defaults of $0.9$ and $0.99$ \citep{yao2021adahessian}. 

\begin{table}[t]
\centering
\caption{Quantitative comparison between the proposed method and the state-of-the-art baseline methods in open-loop settings using the nuScenes dataset. A lower L$2$ error indicates better performance.}
\label{Open-loop-results}
\resizebox{10cm}{!}{%
\begin{tabular}{lcccc}
\toprule
Method & \multicolumn{4}{c}{L2 (m)} \\
\cmidrule{2-5}
& 1s & 2s & 3s & Avg \\
\midrule
NMP \citep{zeng2019end} & - & - & 3.18 & - \\
FF \citep{Hu_2021_CVPR} & 0.55 & 1.20 & 2.54 & 1.43 \\
ST-P3 \citep{hu2022st} & 1.33 & 2.11 & 2.90 & 2.11 \\
UniAD \citep{hu2023planning} & 0.48 & 0.96 & 1.65 & 1.03 \\
VAD-Base \citep{jiang2023vad} & 0.41 & 0.70 & 1.05 & 0.72 \\
\midrule
\textbf{Ours} & \textbf{0.35} & \textbf{0.61} & \textbf{1.01} & \textbf{0.66} \\
\bottomrule
\end{tabular}%
}
\end{table}

\begin{table}[t]
\centering
\caption{Comparison of proposed method with state-of-the-art methods on Town05 Long benchmark in terms of driving score (DS), route completion (RC) and infraction score (IS). $^\ast$ indicates the respective method reports the score on normal all weather conditions, and $^\dagger$ corresponds to adversarial all weather conditions.}
\label{town05}
\resizebox{14cm}{!}{%
\begin{tabular}{@{}lcccc@{}}
\toprule
\textbf{Methods} & \multicolumn{3}{c}{\textbf{Metrics}} \\
\cmidrule(lr){2-4}
& \textbf{DS $\uparrow$} & \textbf{RC $\uparrow$} & \textbf{IS $\uparrow$} \\ 
\midrule
CILRS \citep{codevilla2019exploring,jia2023think}       & $7.800\pm0.30$  & $10.3\pm0.0$  & $0.75\pm0.05$ \\
LBC \citep{chen2020learning,jia2023think}        & $12.30\pm2.00$ & $31.9\pm2.2$  & $0.66\pm0.02$ \\
Transfuser \citep{prakash2021multi,jia2023think}  & $31.00\pm3.60$ & $47.5\pm5.3$  & $0.77\pm0.04$ \\
SDC$^\ast$ \citep{natan2022end}         & $47.13\pm5.27$ & $77.42$  & $0.65$ \\
SDC$^\dagger$ \citep{natan2022end}         & $31.05\pm2.70$ & $64.13$  & $ 0.53$ \\
Roach \citep{zhang2021end,jia2023think}      & $41.60\pm1.80$ & $96.4\pm2.1$  & $0.43\pm0.03$ \\
LAV \citep{chen2022learning,jia2023think}        & $46.50\pm2.30$ & $69.8\pm2.3$  & $0.73\pm0.02$ \\
Think Twice \citep{jia2023think} & $65.00\pm1.70$ & $95.5\pm2.0$  & $0.69\pm0.05$ \\ 
\midrule
\textbf{Ours}        & \bm{$68.30\pm1.90$} & \bm{$96.5\pm1.18$} & \bm{$0.75\pm0.05$} \\ 
\bottomrule
\end{tabular}%
}
\end{table}

\begin{table}[t]
\centering
\caption{Comparison of proposed method with state-of-the-art methods on Longest6 benchmark in terms of driving score (DS), route completion (RC) and infraction score (IS).}
\label{long06-results}
\resizebox{14cm}{!}{%
\begin{tabular}{@{}lcccc@{}}
\toprule
\textbf{Methods} & \multicolumn{3}{c}{\textbf{Metrics}} \\
\cmidrule(lr){2-4}
& \textbf{DS $\uparrow$} & \textbf{RC $\uparrow$} & \textbf{IS $\uparrow$} \\ 
\midrule
LAV \citep{chen2022learning,zhang2023coaching}        & $48.41\pm3.40$ & $80.71\pm0.84$ & $0.60\pm0.04$ \\
Transfuser \citep{prakash2021multi,zhang2023coaching} & $46.20\pm2.57$ & $83.61\pm1.16$ & $0.57\pm0.00$ \\
NEAT \cite{chitta2021neat}\cite{zhang2023coaching}      & $24.08\pm3.30$ & $59.94\pm0.50$ & $0.49\pm0.02$ \\
CAT \cite{zhang2023coaching}       & $58.36\pm2.24$ & $78.79\pm1.50$ & $0.77\pm0.02$ \\
Think Twice \cite{jia2023think}  & $66.7$         & $77.2$         & $0.84$       \\
\midrule
\textbf{Ours}       & \bm{$67.43\pm2.3$}  & \bm{$80.54\pm1.5$}  & \bm{$0.81\pm0.05$} \\
\bottomrule
\end{tabular}%
}
\end{table}

\subsection{Results} \label{Results}
\noindent \textbf{Open-loop Experimental Results on nuScenes:}
The proposed method is evaluated on the L$2$ evaluation metric against the state-of-the-art methods for the quantitative analysis, as illustrated in Table \ref{Open-loop-results}. In our experiments, we have deactivated the ego-status information in the open-loop settings and also fixed the planning horizon $T=3.0s$ to make a fair comparison with the state-of-the-art methods. Since the L$2$ error corresponds to the displacement error in meters between the predicted and ground-truth trajectories, the lower the displacement error, the better the model. The proposed method illustrates better performance as compared to state-of-the-art methods. The comparative analysis uses camera-centric and Lidar-based end-to-end learning methods to predict trajectory. For instance, NMP uses the Lidar and HD maps for predicting future trajectories, giving the L$2$ error of $3.18m$. Then NMP model is only evaluated for the planning horizon of $3.0s$. Similarly, the FF method predicts the future trajectory based on free-space estimation having the L$2$ error of $2.54m$ at the planning horizon of $3.0s$ and an average L$2$ error of $1.43m$. The proposed method illustrates lower L$2$ error at the planning horizon of $3.0s$ and on average compared to NMP and FF methods. Similar to our work, the baseline methods that follow the BEV representation are ST-P$3$, UniAD, and VAD-Base. The L$2$ error for the ST-P$3$, UniAD, and VAD-Base are $2.90m$, $1.65m$, and $1.05m$, respectively, at the planning horizon of $3.0s$, where the proposed method has L$2$ error of $1.01m$ at the same planning horizon, outperforming the ST-P$3$, UniAD, and VAD-Base by $89.5\%$, $38.8\%$, and $3.8\%$ respectively. Similarly, on average, the proposed method shows lower L$2$ error than the state-of-the-art methods. 
\par
\begin{figure*}[t]
      \centering
      \includegraphics[width=14cm]{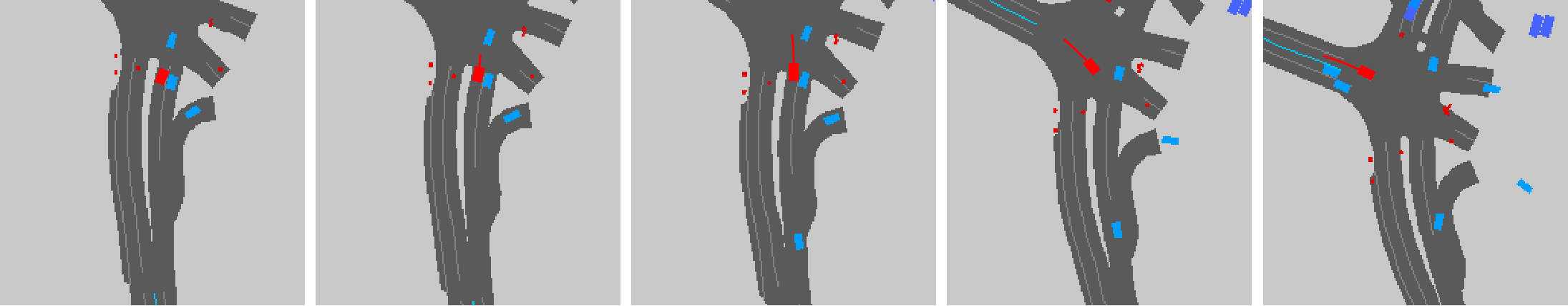}
      \caption{Qualitative results for the proposed method in different driving conditions using nuScenes dataset in open-loop evaluation.}
      \label{Qualitative-nuscenes}
\end{figure*}

Figure \ref{Qualitative-nuscenes} illustrates the qualitative analysis of the proposed method when evaluated on the nuScenes dataset. 

\begin{figure*}[t]
    \centering
    \begin{subfigure}[b]{0.3\textwidth}
        \includegraphics[width=\textwidth]{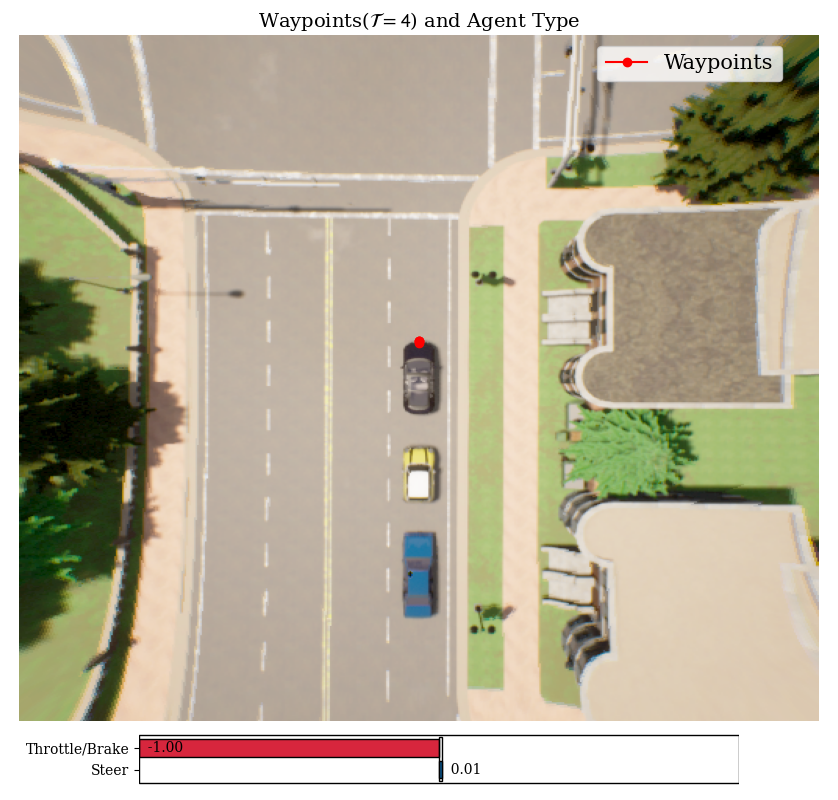}
        \caption{}
        \label{fig:image1}
    \end{subfigure}
    \hfill
    \begin{subfigure}[b]{0.3\textwidth}
        \includegraphics[width=\textwidth]{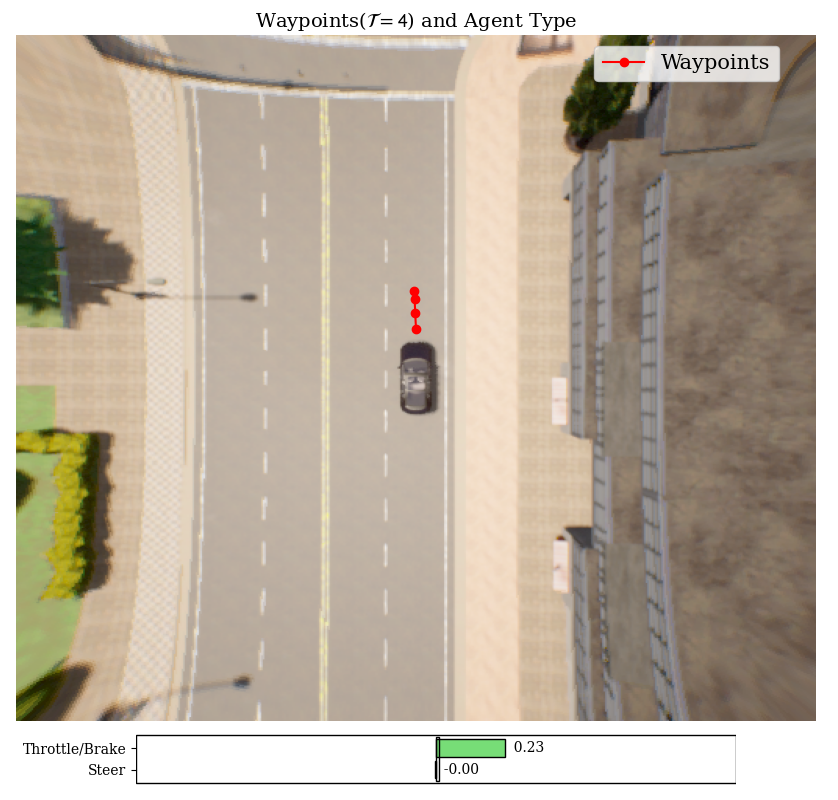}
        \caption{}
        \label{fig:image2}
    \end{subfigure}
    \hfill
    \begin{subfigure}[b]{0.3\textwidth}
        \includegraphics[width=\textwidth]{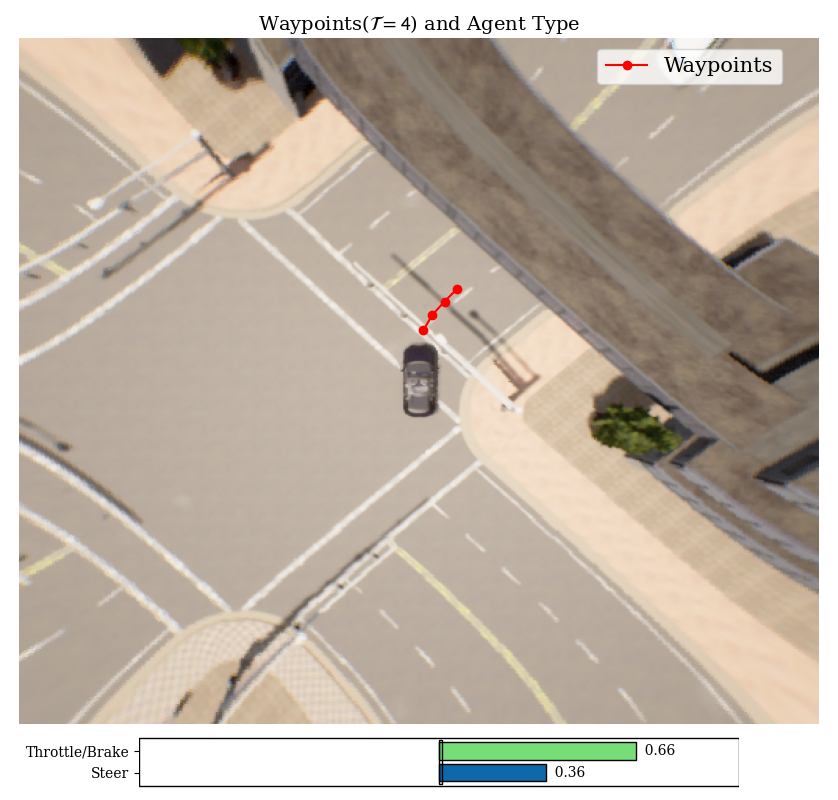}
        \caption{}
        \label{fig:image3}
    \end{subfigure}
    
    \vspace{0.5em}
    
    \begin{subfigure}[b]{0.3\textwidth}
        \includegraphics[width=\textwidth]{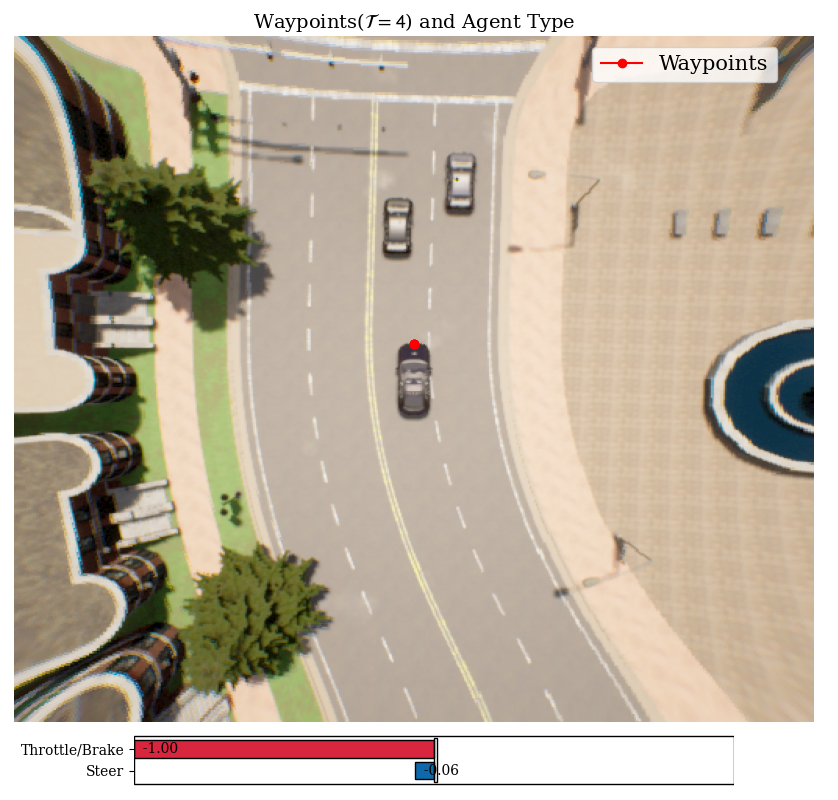}
        \caption{}
        \label{fig:image4}
    \end{subfigure}
    \hfill
    \begin{subfigure}[b]{0.3\textwidth}
        \includegraphics[width=\textwidth]{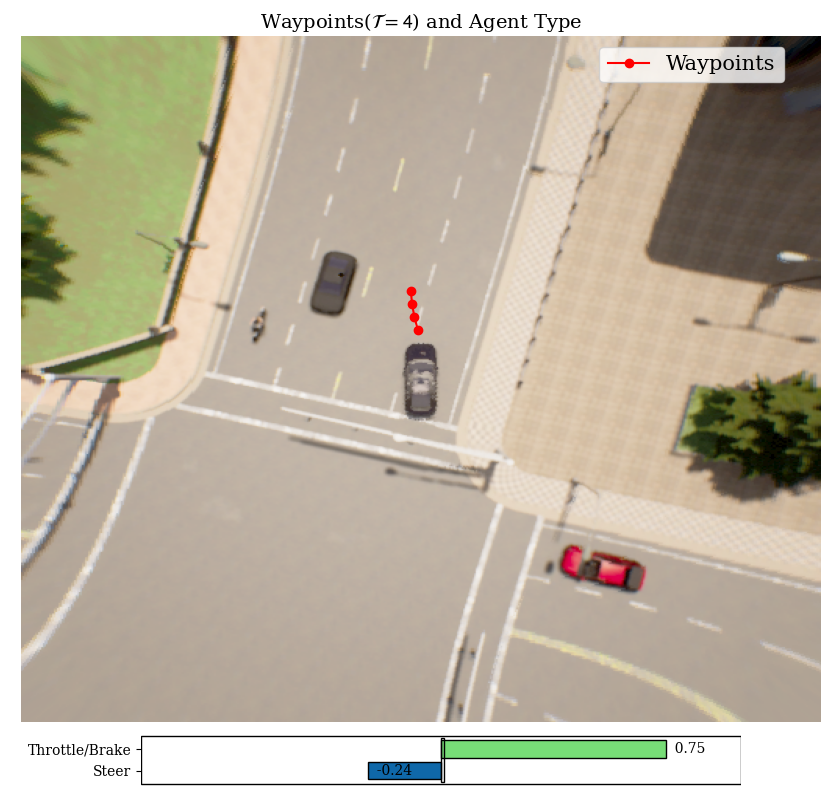}
        \caption{}
        \label{fig:image5}
    \end{subfigure}
    \hfill
    \begin{subfigure}[b]{0.3\textwidth}
        \includegraphics[width=\textwidth]{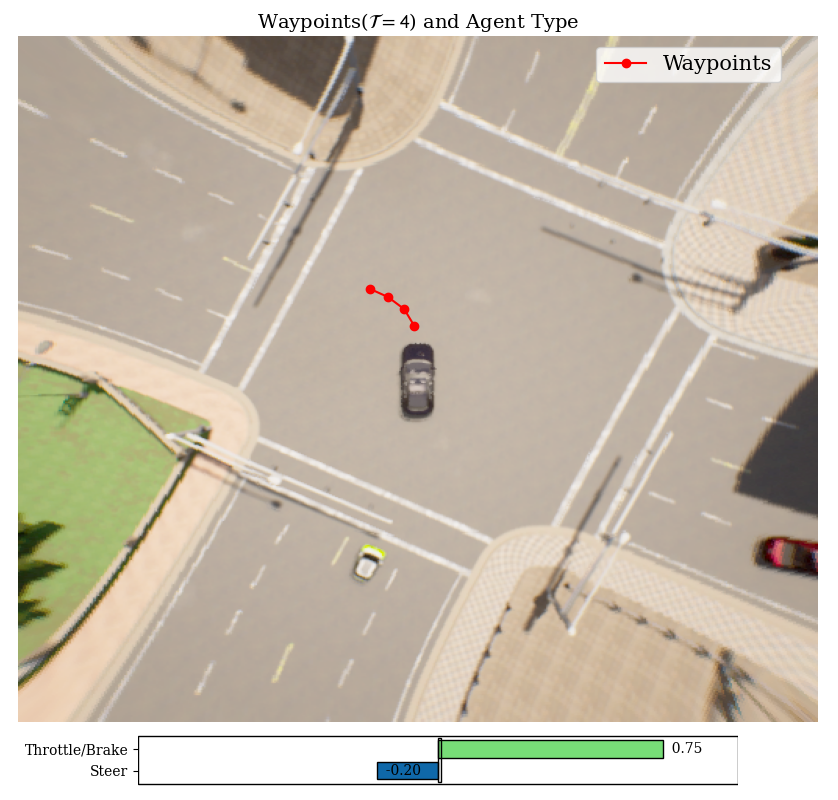}
        \caption{}
        \label{fig:image6}
    \end{subfigure}
    \caption{Qualitative results for the proposed method in different driving conditions on Town05 Long benchmark.}
    \label{town05results}
\end{figure*}

\begin{figure*}[t]
    \centering
    \begin{subfigure}[b]{0.3\textwidth}
        \includegraphics[width=\textwidth]{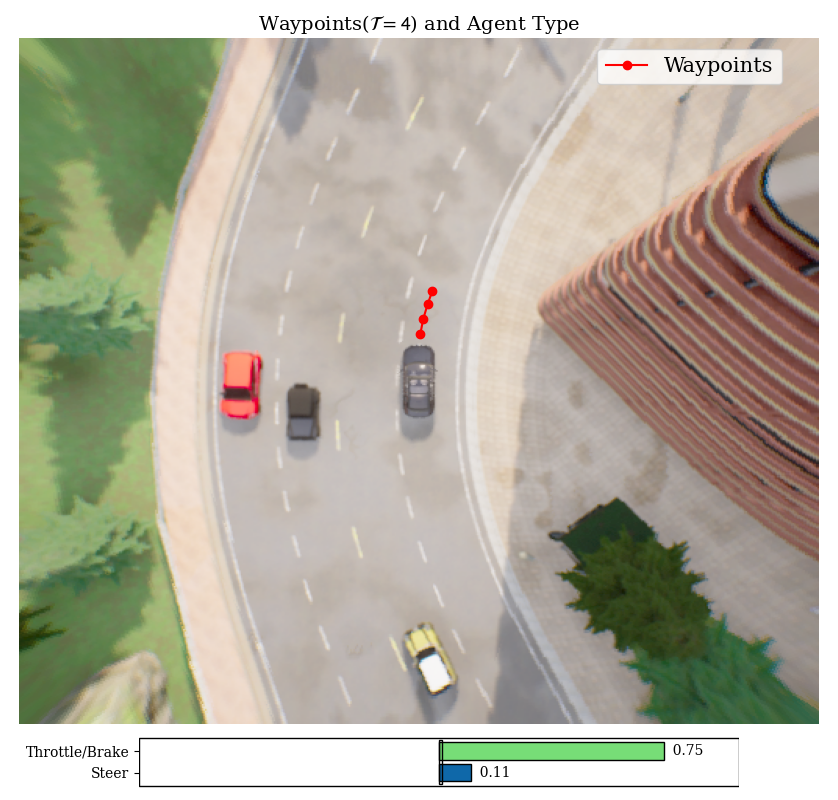}
        \caption{}
        \label{fig:image1}
    \end{subfigure}
    \hfill
    \begin{subfigure}[b]{0.3\textwidth}
        \includegraphics[width=\textwidth]{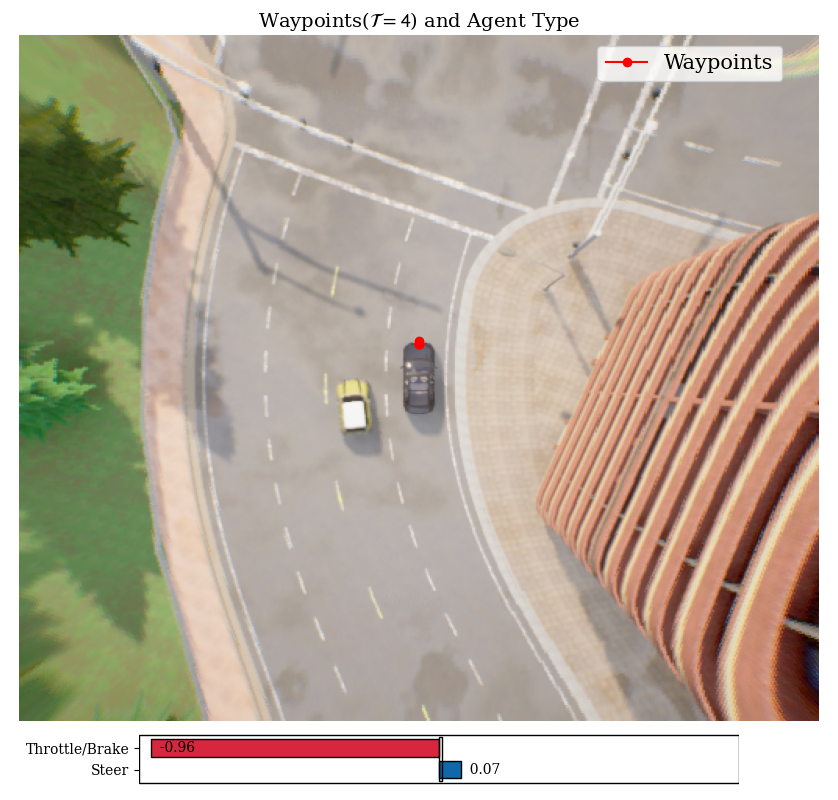}
        \caption{}
        \label{fig:image2}
    \end{subfigure}
    \hfill
    \begin{subfigure}[b]{0.3\textwidth}
        \includegraphics[width=\textwidth]{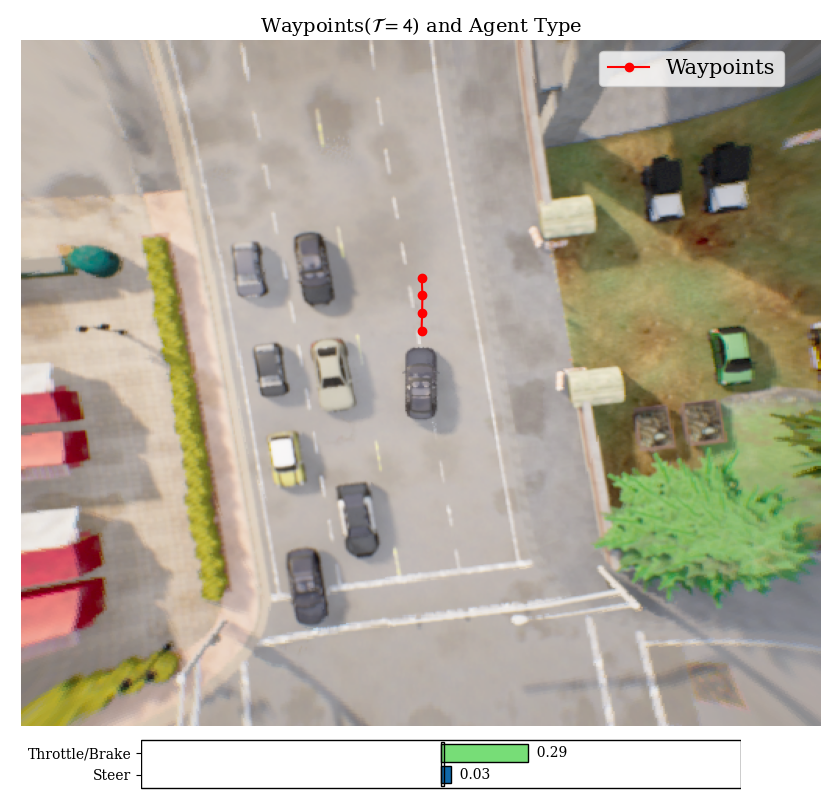}
        \caption{}
        \label{fig:image3}
    \end{subfigure}
    
    \vspace{0.5em}
    
    \begin{subfigure}[b]{0.3\textwidth}
        \includegraphics[width=\textwidth]{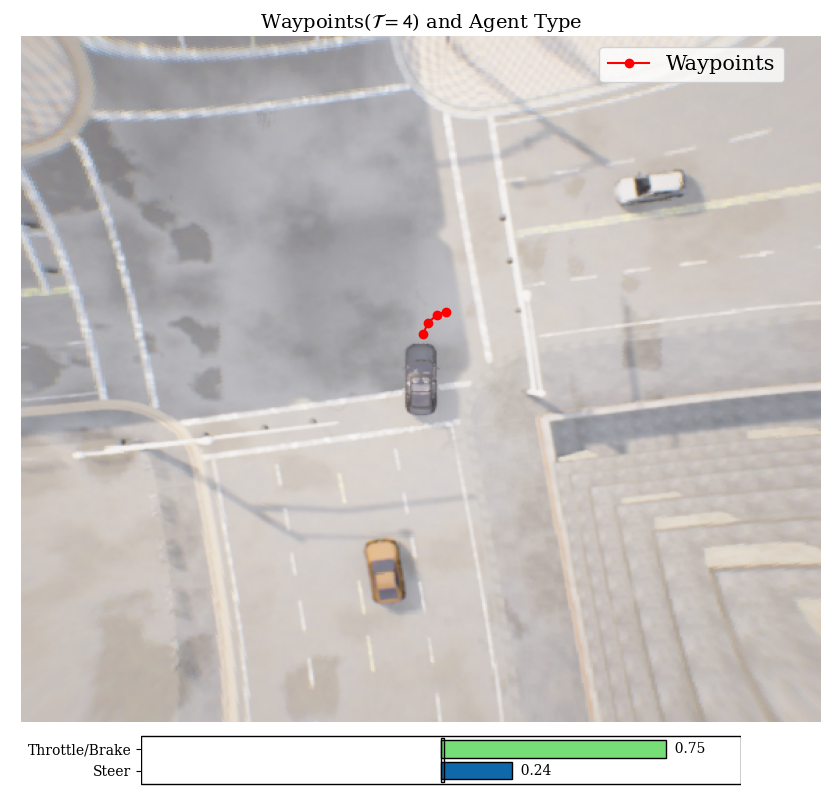}
        \caption{}
        \label{fig:image4}
    \end{subfigure}
    \hfill
    \begin{subfigure}[b]{0.3\textwidth}
        \includegraphics[width=\textwidth]{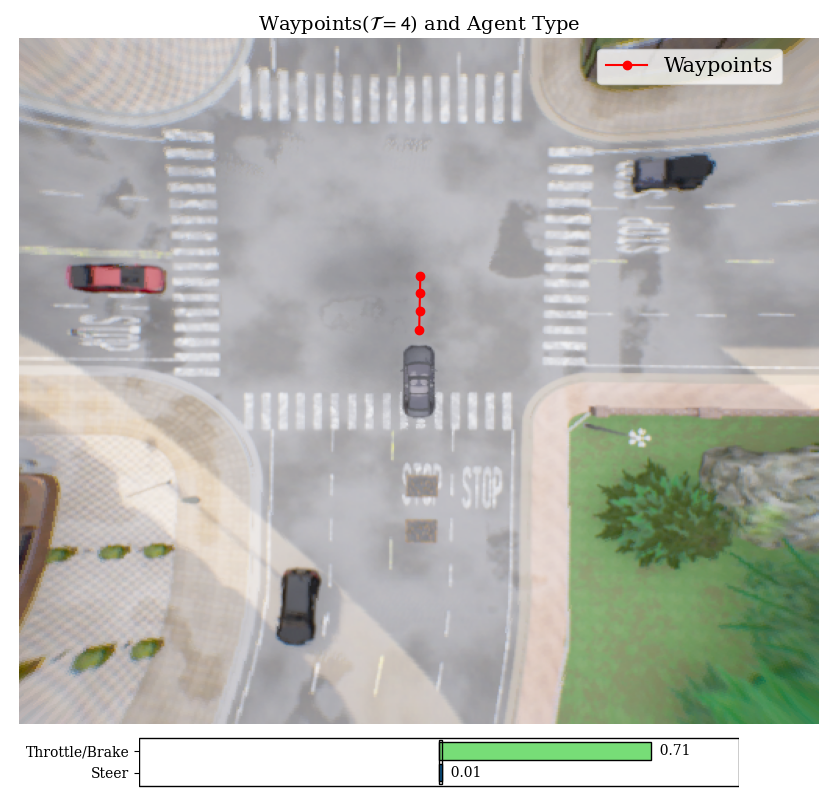}
        \caption{}
        \label{fig:image5}
    \end{subfigure}
    \hfill
    \begin{subfigure}[b]{0.3\textwidth}
        \includegraphics[width=\textwidth]{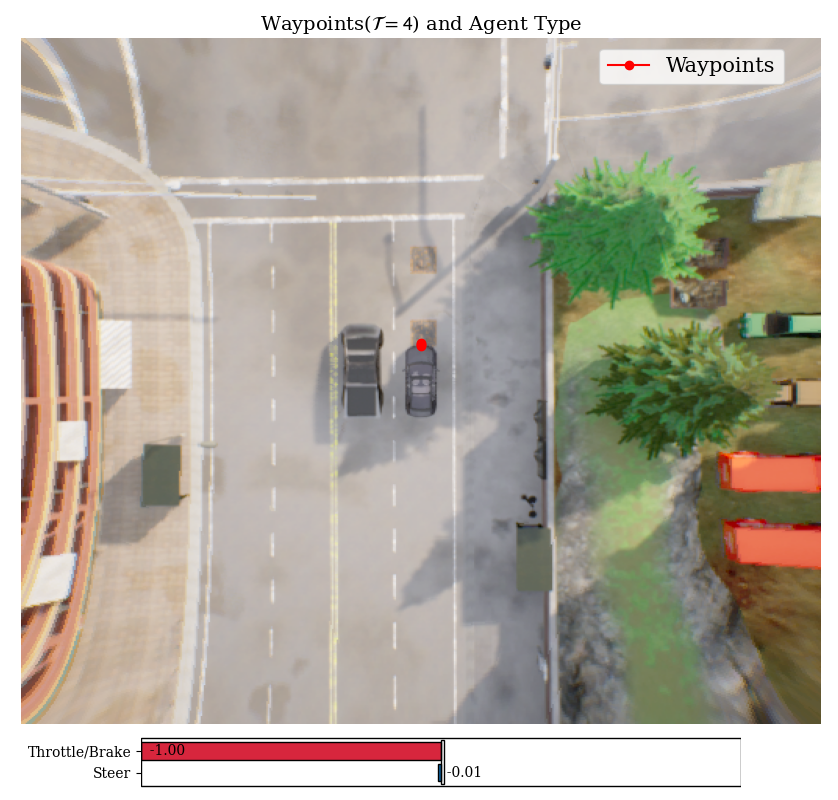}
        \caption{}
        \label{fig:image6}
    \end{subfigure}
    \caption{Qualitative results for the proposed method in different driving conditions on Longest6 benchmark.}
    \label{long06results}
\end{figure*}

\noindent \textbf{Closed-loop Experimental Results on CARLA:}
We compare the proposed method with other state-of-the-art methods on two CARLA benchmarks, Town05 Long and Longest6, in closed-loop settings. Our quantitative analysis considers the baselines with multi-modality inputs rather than sticking with methods involving only a single modality. Using contextual information, the proposed method achieves better driving, route completion, and infraction scores. Table \ref{town05} illustrates the quantitative results of the proposed method on the Town05 Long benchmark. Specifically, the proposed method achieves the driving score of $68.30 \pm 1.90$, $96.5 \pm 1.18$ of route completion, and $0.75\pm0.05$ of infraction score, outperforming the ThinkTwice by $4.8\%$ in driving score, $1.03\%$ in route completion and $8\%$ in infraction score respectively. Similarly, the proposed method illustrates better evaluation metrics scores when compared with other state-of-the-art methods. 
\par
Table \ref{long06-results} shows the proposed method results with other state-of-the-art methods on the Longest6 benchmark in closed-loop settings. The proposed method achieves the driving score of $67.43\pm2.3$, $80.54\pm1.5$ of route completion, and $0.81\pm0.05$ of infraction scores on the Longest6 benchmark, outperforming the other state-of-the-art methods in evaluation metrics in the closed-loop settings. Figure \ref{town05results} and \ref{long06results} illustrate the proposed method's qualitative results on Town05 and Longest6 benchmarks in various driving scenarios.The learned driving policy through the proposed method is displayed in moving straight, stopping at the traffic light, and making left, and right turns. These results demonstrate that the driving policy learned using the proposed method show promising results and complements the quantitative analysis of the proposed method with other state-of-the-art baseline methods.

\section{Conclusion} \label{Conclusion}
In this work, we explore the use of contextual information for learning driving policies in an end-to-end manner for autonomous driving. Drawing inspiration from the human neural map representation of the environment, we employ three RGB cameras coupled with a top-down semantic map to achieve a holistic understanding of the surroundings. This environmental representation is then channeled through a self-attention-based perception module, subsequently processed by a GRU-based waypoint prediction module for generating the waypoints. The proposed method is experimentally evaluated for both open-loop and closed-loop settings, illustrating better performance than state-of-the-art methods. 
\par
Building on this foundation, there are avenues for further exploration. While our current framework primarily relies on RGB cameras and a semantic map for environmental perception, future research could benefit from incorporating additional sensors, such as radar and LiDAR, to enhance the perception module. An intriguing area of investigation remains: how to refine the contextual representation of the environment for driving policy predictions, especially when integrated with neural-network-based controllers. This presents a promising direction for advancing the capabilities of autonomous driving systems.

\section*{Acknowledgments}

This work is supported by the Academy of Finland
Flagship program: Finnish Center for Artificial Intelligence (FCAI), and also by Culture, Sports and Tourism Research and Development Program through the Korea Creative Content Agency Grant funded by the Ministry of Culture, Sports
and Tourism (R2022060001, “Development of Service Robot and Contents Supporting Children’s Reading Activities Based on Artificial Intelligence”).



\newpage
\bibliographystyle{elsarticle-harv} 
\bibliography{cas-refs}





\end{document}